\documentclass{article}

\usepackage[final]{arxiv}

\usepackage{subfig}
\usepackage{times}
\usepackage{epsfig}
\usepackage{graphicx}
\usepackage{amsmath}
\usepackage{amssymb}
\usepackage{caption}
\usepackage{multirow}
\usepackage{color}
\usepackage{threeparttable}
\usepackage{tabularx}
\usepackage{siunitx}
\usepackage{float}
\usepackage{wrapfig}

\usepackage{hyperref}

\title{End-to-End Learning of Semantic Grasping}

\author{
  Eric Jang\\
  Google Brain\\
  \texttt{ejang@google.com} \\
  \And
  Sudheendra Vijayanarasimhan \\
  Google \\
  \texttt{svnaras@google.com} \\
  \And
  Peter Pastor \\
  X \\
  \texttt{peterpastor@x.team} \\
  \AND
  Julian Ibarz \\
  Google Brain \\
  \texttt{julianibarz@google.com} \\
  \And
  Sergey Levine \\
  Google Brain, UC Berkeley \\
  \texttt{slevine@google.com}
}

\begin{document}
\maketitle


\begin{abstract}
We consider the task of semantic robotic grasping, in which a robot picks up an object of a user-specified class using only monocular images. Inspired by the two-stream hypothesis of visual reasoning, we present a semantic grasping framework that learns object detection, classification, and grasp planning in an end-to-end fashion. A ``ventral stream'' recognizes object class while a ``dorsal stream'' simultaneously interprets the geometric relationships necessary to execute successful grasps. We leverage the autonomous data collection capabilities of robots to obtain a large self-supervised dataset for training the dorsal stream, and use semi-supervised label propagation to train the ventral stream with only a modest amount of human supervision. We experimentally show that our approach improves upon grasping systems whose components are not learned end-to-end, including a baseline method that uses bounding box detection. Furthermore, we show that jointly training our model with auxiliary data consisting of non-semantic grasping data, as well as semantically labeled images without grasp actions, has the potential to substantially improve semantic grasping performance\footnote{A video overview of this paper can be found at \url{https://youtu.be/WR5WUKXUQ8U}.}.
\end{abstract}



\section{Introduction}
\label{sec:introduction}

Robotic manipulation presents a number of challenges as well as exciting opportunities for computer vision. Robots that interact with the world must recognize and localize objects with a very high degree of precision, which places a high demand on the performance of computer vision systems. The substantial domain shift between objects and scenes seen by a robot and those in labeled object recognition datasets makes it nontrivial to generalize conventional computer vision systems to robotic tasks. On the other hand, robots have the ability to autonomously collect their own data as they perform manipulation tasks in the real world. By developing methods for self-supervision, we can utilize robots to autonomously collect large self-labeled datasets that can help us exceed the performance of standard vision systems that are trained on fixed-size, human-labeled datasets.

In this paper, we explore this in the context of robotic grasping with semantic labels. The purpose of our method is to allow a robot to successfully grasp objects in its environment based on high-level user commands, which specify the type of object to pick up. This task aims to combine both spatial and semantic reasoning: the robot must determine which of the objects in the scene belong to the requested class, and how those objects might be grasped successfully. We aim to answer two questions: first, how can we design a model that effectively combines both spatial and semantic reasoning, and second, how can we exploit the capacity of the robots to collect self-supervised data while also making use of modest numbers of human-provided labels to learn about semantics. Answering these questions successfully could lead not only to more effective and useful robot systems, but to advances in spatial and semantic visual reasoning and automated data gathering.

Previous work~\cite{levine16, lg-sss-15} has demonstrated that a non-semantic grasp outcome predictor can be trained in an entirely self-supervised manner via autonomous data collection. However, those works did not consider object class in grasp planning. In order to solve the semantic grasping task, we require semantically labeled data and a model that is capable of reasoning about object identity.

\begin{figure}[t!]
\centering
\includegraphics[width=\linewidth]{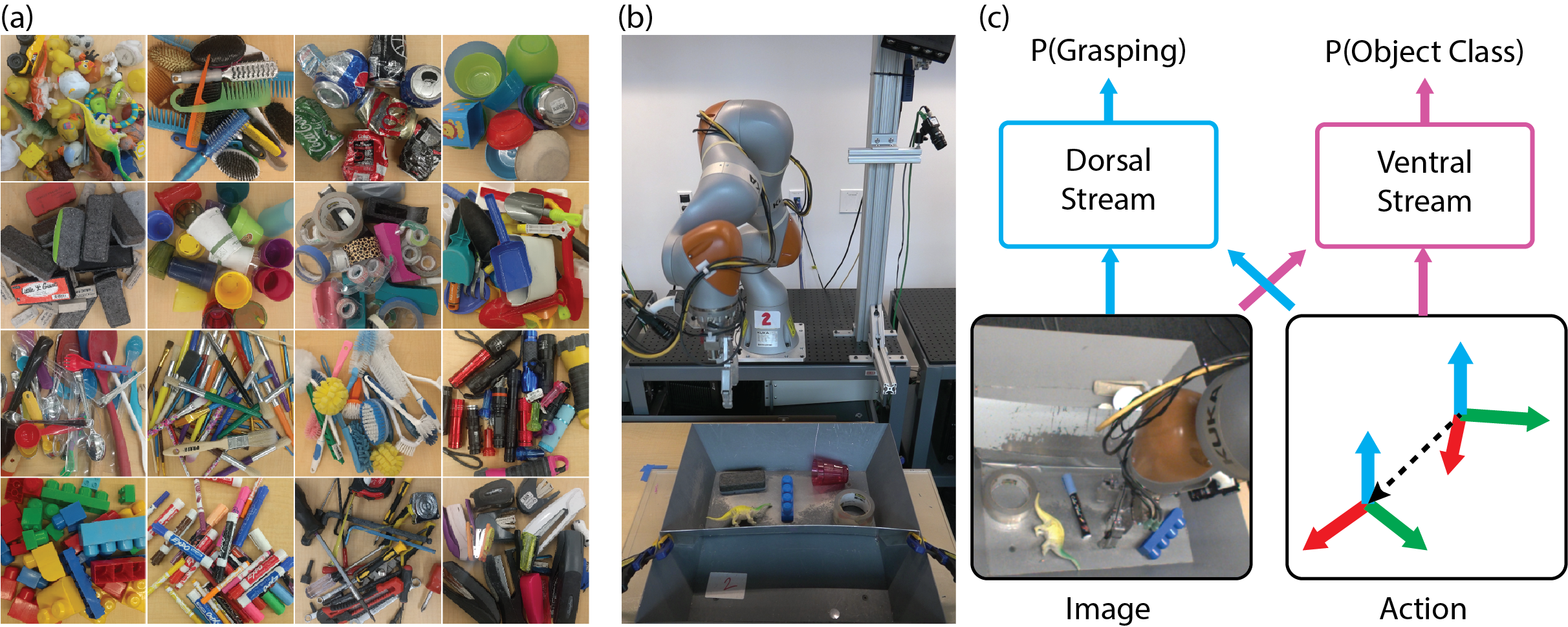}
\caption{(a) We consider the task of learning to pick up objects from 16 object classes, conditioned on a user-specified class. (b) Our robotic setup is an arm with a 2-finger gripper that servos using monocular camera images. (c) Inspired by the two-stream hypothesis of vision \cite{uh-wwhb-94}, we propose a model that is divided into a dorsal and ventral stream: the dorsal stream selects which actions result in good grasps, while the ventral stream selects the class of the object being grasped. Best viewed in color.}
\label{fig:model}
\end{figure}

In designing our method, we take inspiration from the ``two-stream hypothesis'' of human vision, where visual reasoning that is both spatial and semantic can be divided into two streams: a ventral stream that reasons about object identity, and a dorsal stream that reasons about spatial relationships without regard for semantics \cite{uh-wwhb-94}. We train a deep neural network for performing the semantic grasping task that is decomposed into these two streams, as illustrated in Figure~\ref{fig:model}. The dorsal stream predicts whether a particular grasping action will result in a successful grasp. The ventral stream is trained to predict what type of object will be picked up. We tackle the problem of assigning semantic labels to large amounts of robot-collected data via a label propagation scheme: if the robots successfully grasp an object, they can then bring it up to their cameras for closer inspection. A modest number of such human-labeled images can be used to train a classifier for these images, and then propagate labels to the images used for learning grasping in a semi-supervised fashion. 

The main contribution of our work is a two-stream neural network model for learning semantic grasping from off-policy data. We use a combination of self-supervised robotic data collection and label propagation to construct large object classification dataset for training our models. Finally, we show that both dorsal and ventral streams can benefit independently from domain transfer, by integrating auxiliary spatial information from non-semantic grasping datasets and auxiliary semantic information from image classification datasets. 

Our experimental results demonstrate that: (1) Our proposed two-stream model substantially outperforms a more standard detection pipeline for grasping, as well as “single-stream” approaches. (2) Integrating auxiliary non-semantic data substantially improves grasping performance. (3) Auxiliary semantically labeled data without grasping actions has the potential to improve generalization, but this improvement is highly dependent on the degree of domain shift between the auxiliary data and the grasping images.

\section{Related Work}
\label{sec:related}

We consider semantic grasping as our model task, where the robot must pick up objects of a specific, user-selected semantic category. A complete discussion of prior grasping work is outside of the scope of this paper, and a more complete overview can be found in recent survey work \cite{bohg2014}. In contrast to most grasping methods in robotics \cite{goldfeder2009,weisz2012,rodriguez2012}, we use only monocular RGB images, and do not make use of any hand-designed geometric cues, camera calibration, or even feedback from the gripper to execute a grasp sequence. Prior data-driven methods in grasping have used human supervision \cite{herzog2014,lenz2015} and automated geometric criteria \cite{goldfeder2009grasp_database} for supervision. Both types of data-driven grasp selection have recently incorporated deep learning \cite{kappler2015,lenz2015,redmon2015,rob_platt_paper}.

While most grasping methods focus on the spatial aspects of the problem, a number of works have sought to incorporate semantic information into grasping, including affordances and task-specific knowledge \cite{dang2014,katz2014,gupta2015}, instance-based object models \cite{oberlin2015}, known shape information about objects \cite{goldfeder2009grasp_database}, and prior knowledge that objects of the same type should be grasped in similar ways \cite{nikandrova2015}. However, while these prior methods incorporate semantic knowledge to aid spatial reasoning~\cite{kehoe2014}, our approach instead aims to combine both spatial and semantic reasoning to pick up objects of a particular category in cluttered scenes. Incorporating additional feedback from semantic knowledge into the spatial problem of grasping, as in prior work, is also a very valuable research direction, but largely orthogonal to the main focus of this work. In this sense, our method is complementary.

Self-supervision and large-scale data collection have recently been proposed for image-based robotic grasping \cite{lg-sss-15,levine16}. Our work also follows this paradigm for training the dorsal stream of our semantic grasping network, with large numbers of grasp attempts collected with real robotic manipulators and labeled as successful or failed grasps automatically. However, we also incorporate a ventral stream into our system that is trained on semantic categories, with knowledge transfer from object classification datasets and automatic label propagation from a small number of human-provided labels. In this sense, our aims are orthogonal to this prior work; our goal is not to simply build a better grasping system, but to explore how semantics and label propagation interact with real-world robotic manipulation tasks.

\section{Deep Semantic Grasping}
\label{sec:approach}

In our setup, a robotic arm with a two-finger gripper learns to grasp particular objects from a bin based on a user-provided object class. We train our grasping system end-to-end, using off-policy data collected by a non-semantic grasping policy and a modest number of semantic labels provided by humans. We propose several deep neural net architectures for learning semantic grasping, discuss our methodology for collecting large datasets for training our models, and describe how information from external datasets can be leveraged to improve grasping performance.

\subsection{Two-Stream Value Function}
\label{sec:twostream_arch}

\begin{figure*}[t!]
\centering
\includegraphics[width=0.9\linewidth]{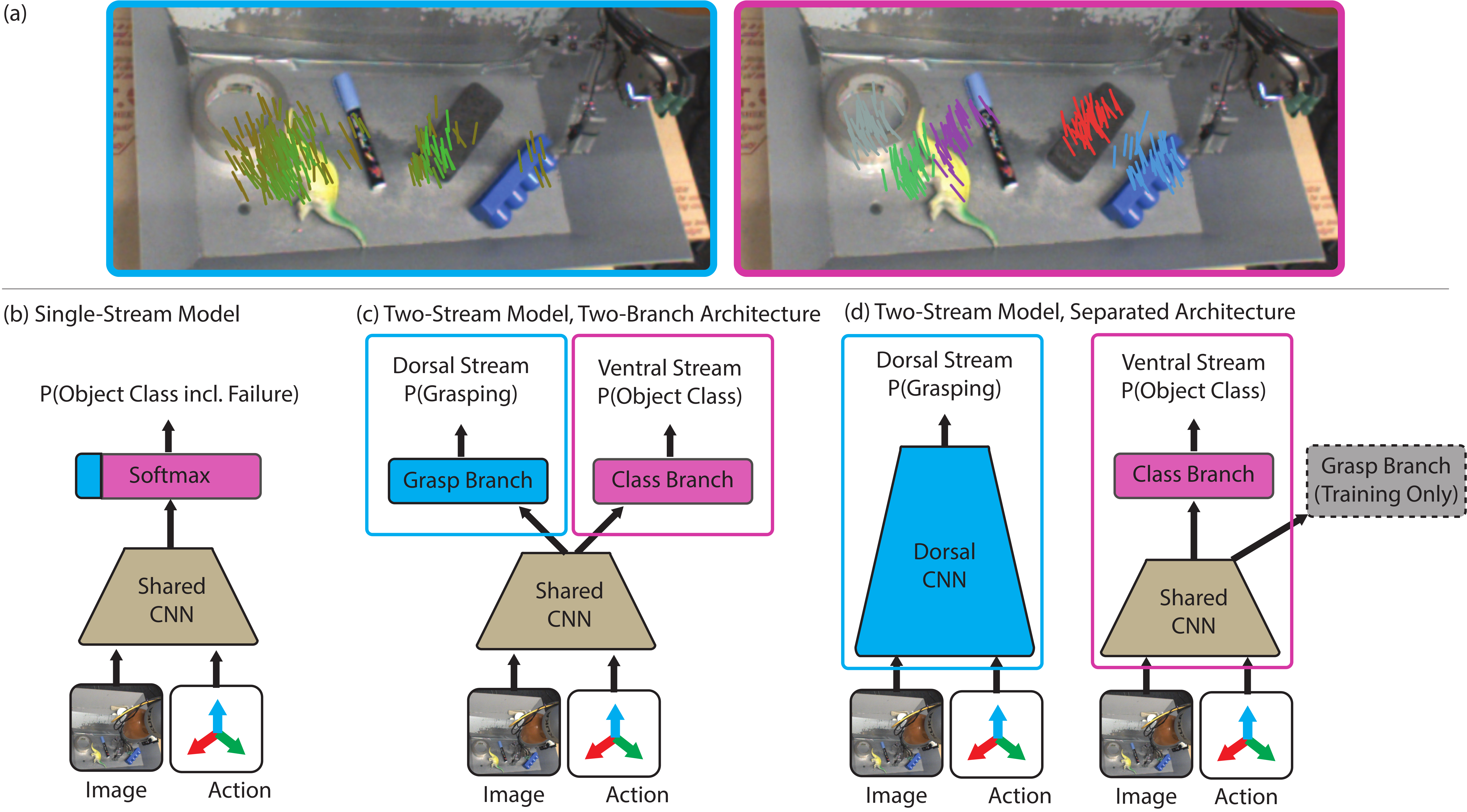}
\caption{(a) Left: visualization of grasp actions projected into screen space, with brighter green actions corresponding to actions with higher probability of grasp success. Right: projected grasp actions, with each color corresponding to the most probable actions for picking up an object class. Hand-eye coordination binds spatial awareness with visual semantics, enabling the policy to learn object classification and detection end-to-end. (b) Single Stream model that predicts grasp success for each class independently, without explicitly modeling object semantics. (c) Two-Stream model that shares model parameters between a grasp branch and class branch, which comprise the dorsal (cyan box) and ventral streams (magenta box). (d) Our best two-stream model uses the class branch from (c) for the ventral stream, but uses a separate dorsal stream trained only on grasp prediction. Best viewed in color.}
\label{fig:twostream_family}
\end{figure*}

Our policy critic, visualized in Figure~\ref{fig:model}c, takes as input images $I_t, I_0$, which correspond to the current image seen by the robot's camera, and the initial image seen during the episode. The initial image is recorded before the grasping episode begins and is included to allow the system to handle self-occlusions caused by the robot arm. The network also receives as input a candidate task-space motion command $a_t$, which is parameterized by a Cartesian displacement of the gripper along with a vertical rotation of the gripper. The network outputs $p(g, c | I_t, a_t)$, the joint probability that executing the command $a_t$ will result in $g$, the event that an object is grasped and $c$, the event that the grasped object is the correct semantic label. Our servoing function samples grasps randomly and uses the neural critic to choose the grasp with the highest probability of picking up the desired object. See Appendix~\ref{sec:experiment_details} for additional details on the model architecture and the servoing policy.

A straightforward way to implement a semantic neural critic is to extend the final layer of the model proposed in prior work on self-supervised grasping~\cite{levine16} to output a softmax distribution over object classes, plus an additional ``failure'' class for when no objects are grasped (Figure~\ref{fig:twostream_family}b). This model, which we call ``Single Stream'' in our experiments, lacks a dorsal stream that explicitly learns what it means to ``pick up any object''. Instead, it merges spatial and semantic inference into a single output, predicting grasp success for each object class independently.

Additional supervision of spatial reasoning capabilities can be obtained by distinguishing between failures of classification and failures of detection. We decompose the prediction problem into a dorsal stream $p(g|I_t,a_t)$ that is trained on non-semantic grasping data (whose labels are success or failure), while the ventral stream $p(c|I_t,a_t,g)$  is trained to classify successful grasps whose labels are one of 16 object classes. The model learns to infer where actions will send the gripper, and implicitly attends to that location in order to extract its spatial relevance to grasping (Figure~\ref{fig:twostream_family}a, left) and semantic identity (Figure~\ref{fig:twostream_family}a, right).

We consider Two-Stream models as not a single neural net architecture, but rather a family of architectures that functionally decouple the prediction of grasp success (dorsal stream) and object class (ventral stream). An example of a Two-Stream model is a ``Two-Branch'' (2B) architecture that shares inputs and action processing before branching out to grasp-specific and class-specific parameters (Figure~\ref{fig:twostream_family}c). Another example is a ``Separated'' architecture which takes advantage of the fact that Two-Stream models can mix-and-match dorsal and ventral streams at inference time. We train both branches of the 2B architecture as before, but only use the class branch as the ventral stream. The dorsal stream of the Separated architecture is an entirely separate convolutional network trained only to predict grasp success (Figure~\ref{fig:networkarch}b). Because this separate dorsal stream outperforms non-semantic grasping capabilities of our 2B models, we use this separated architecture for most of our Two-Stream models. \footnote{On terminology: we use ``Two-Stream model'' to refer to the functional decomposition of spatial and semantic reasoning into the multiplication of $p(g|I_t,a_t) \cdot p(c|I_t,a_t,g)$, while we use ``Two-Branch'' and ``Separated'' to reflect architectural choices on how to implement a Two-Stream model.}. Sharing the dorsal stream between Two-Stream architectures also allows for a controlled comparison between different ventral stream architectures.

Instead of using a deep CNN that implicitly attends to an object using the action vector, we wondered whether an explicit attention mechanism would improve the model's ability to extract a representation of visual features that was invariant to the object's location in the bin. In addition to our 16-layer CNN, we also try a shallower 9-layer network that bottlenecks predictions through soft keypoint locations~\cite{dsae}. Both architectures perform similarly, though the soft keypoint architecture has substantially fewer parameters and therefore trains more efficiently. Further details on both architectures are provided in Appendix~\ref{sec:soft_attn}.

\begin{figure}[t]
\centering
\includegraphics[width=0.9\linewidth]{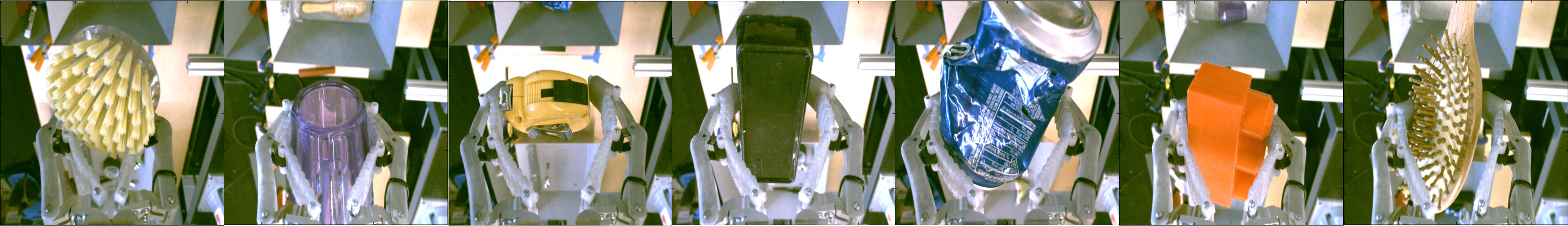}
\caption{Images captured during the execution of the ``present'' pose of the robot, which facilitate easier classification and label propagation to cluttered grasping images.}
\label{fig:wrist}
\end{figure}

\subsection{Parallel Data Collection \& Model Training}

We chose $16$ object categories found in typical kitchen and office settings (Table \ref{table:object_classes}). We obtained 30-40 object instances per category and divided these 500 objects into a set of training and test objects (which are used to evaluate generalization to unseen class instances).
To collect the data used to train our semantic grasping models, we first learn a non-semantic grasping policy that is capable of picking up objects in a self-supervised manner, similar to the approach from~\cite{levine16}. We utilized 8 KUKA iiwa arms to collect data in parallel and introduce diverse lighting and variations in camera calibration into the dataset. Labels for grasp success/failure can be generated autonomously via a simple threshold on an image delta, as described in~\cite{levine16}.

\begin{table}[]
\centering
\resizebox{\linewidth}{!}{
\begin{tabular}{c|c|c|c|c|c|c|c|c|c|c|c|c|c|c|c|c}
\textbf{Object Category} & 
\rotatebox{45}{Eraser} &  
\rotatebox{45}{Scotchtape} & 
\rotatebox{45}{Comb} & 
\rotatebox{45}{Shovel} & 
\rotatebox{45}{Legos} &  
\rotatebox{45}{Torch} &  
\rotatebox{45}{Pen} &  
\rotatebox{45}{Plasticcup} &  
\rotatebox{45}{Toy} &  
\rotatebox{45}{Tools} &
\rotatebox{45}{Stapler} & 
\rotatebox{45}{Toiletbrush} & 
\rotatebox{45}{Paintbrush} & 
\rotatebox{45}{Spoon} & 
\rotatebox{45}{Bowl} & 
\rotatebox{45}{Soda can} \\
\hline
\textbf{Human-rated training examples} & 395 & 464 & 587 & 649 & 591 & 288 & 437 & 517 & 1191 & 189 & 860 & 128 & 613 & 331 & 201 & 91 \\
\textbf{Human-rated test examples} & 14 & 110 & 320 & 4 & 123 & 25 & 60 & 286 & 533 & 62 & 28 & 79 & 114 & - & 312 & - \\
\end{tabular}}
\caption{The object categories in our dataset along with the number of training and testing instances.}
\label{table:object_classes}

\end{table}

Next, we use two approaches to annotate the collected dataset with object labels, in order to supervise learning of the ventral branch. In the first approach, each robot's bin is filled with objects sharing the same semantic label. This makes it trivial to assign labels to large amounts of data, for which we gather 10,000 such examples. Object collections are frequently shuffled between the different robots so that our models do not overfit to a specific robot's idiosyncratic visual cues. In our second approach, we program the robot to lift grasped objects close to the camera at the end of a grasping sequence for futher inspection, at which point a close-up image of the object being held in the gripper is recorded (e.g. Figure~\ref{fig:wrist}). Presenting objects in this manner isolates the grasped objects for the purpose of easier classification, somewhat analogously to how children pick up and examine objects in order to isolate it from surrounding visual context. We label 20,000 present images using Amazon Mechanical Turk and use this information to train a CNN classifier. By classifying the remaining 300,000 present images and propagating those labels to their corresponding grasping images, we can generate a large autonomously collected dataset for semantic grasping.

\subsection{Transfer from External Datasets}
\label{sec:transfer}

Although the robots can gather large amounts of data autonomously and then label them via automatic label propagation, the diversity of object instances in such datasets is limited. Fortunately, our two-stream decomposition of spatial and semantic perception affords us a straightforward mechanism for incorporating images from either (1) auxiliary grasping data with no semantic labels, including objects that are difficult to classify or do not belong to the semantic classes, or (2) auxiliary semantic data of different object instances without grasping actions.

For auxiliary grasping data, we jointly train our two-stream models on all the grasping data we collected, including objects that do not belong to our semantic categories, supervising only the grasp branch of the model. In all, we had 11 million grasping images that we could use to train the dorsal stream (corresponding to about 2 million grasps), of which 330,000 had semantic labels and could be included for training the ventral stream. This auxiliary data serves to not only improve our model's ability to generalize to the test objects, but also benefits classification by improving the early visual representations learned by the shared CNN. To incorporate auxiliary semantic data, we include images from object classification datasets directly into the grasping model with a random action.

In our experiments, we present results with two auxiliary semantic datasets, which we call ``S1'' and ``S2.'' S1 contains images from ImageNet and JFT~\cite{imagenet, hinton2015distilling} and with present images of training objects. S2 contains present images of objects in the test set (but are not paired with grasping actions). The rationale behind including S2 in our evaluations is to determine how much improvement can be had from a classification dataset that perfectly represents the distribution of objects seen at test time, from a different viewpoint and with a different camera. This represents a ``best case'' scenario for semantic but non-spatial transfer. The performance of the method with both of these two datasets is presented in Section~\ref{sec:transfer_experiments}.


\section{Experimental Results}
\label{sec:experiments}

\begin{figure}[t!]
\centering
\includegraphics[width=\linewidth]{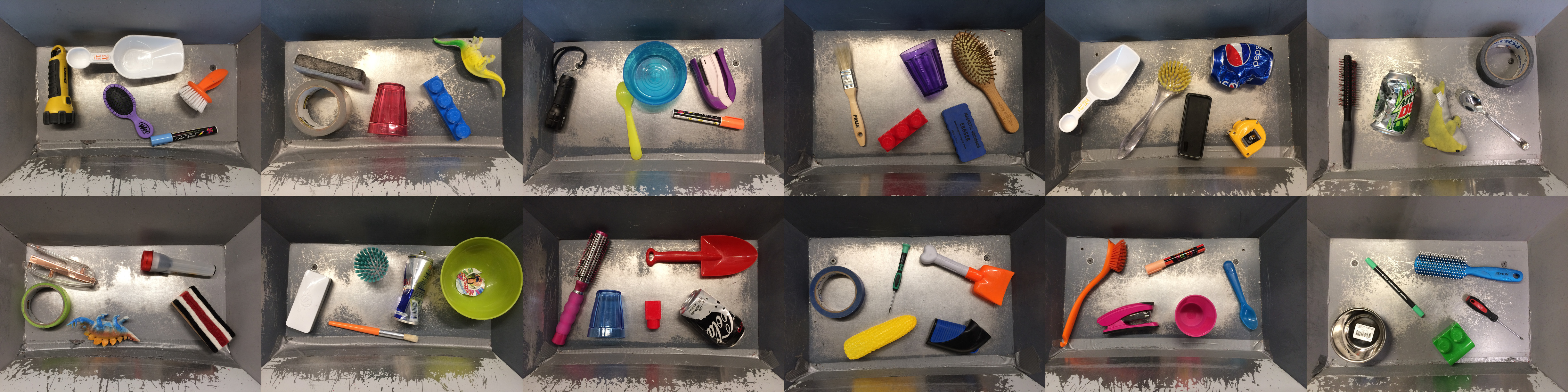}
\caption{In our evaluation, the robots are tasked to pick up specific categories from bins containing 5 randomly selected objects from different classes. Top row: Training objects. Bottom row: Test objects. Evaluation objects were selected at random to cover all object categories uniformly.}
\label{fig:eval_objs}
\end{figure}

We evaluate semantic grasping performance on real-world robots by assigning 5 randomly selected objects from different categories to each robot (Figure \ref{fig:eval_objs}) and commanding the robot to pick up the objects in a random order. For a sequence of 5 grasp attempts, the robot removes grasped objects one at a time from the bin and drops them outside the workspace. This is repeated 10 times per robot with 6 robots in total, for a set of 30 training objects and 30 unseen testing objects. As shown in Appendix~\ref{sec:results_table}, we evaluate policies based on the fraction of grasp attempts that were successful at picking up some object (Objects/Attempts) and the fraction of all grasp attempts that picked up the correct object (Class/Attempts). We demonstrate through baseline comparisons and ablation studies the effect of various architectural decisions in our semantic grasping models.

{\bf Summarily:} (1) End-to-end learning of semantic grasping outperforms a traditional detection-classification-grasping method. (2) Two-stream decompositions of grasp and class prediction outperform a single-stream model. (3) A separated architecture is better at grasping than two-branch architectures. (4) A 9-layer CNN with attention performs comparably to a 16-layer CNN without attention. (5) Auxiliary grasping data improves classification of our two-stream architectures. (6) Auxiliary semantic data that reflects the distribution of objects we are evaluating on improves class grasping accuracy.

\begin{figure}[t!]
\centering
\includegraphics[width=.9\linewidth]{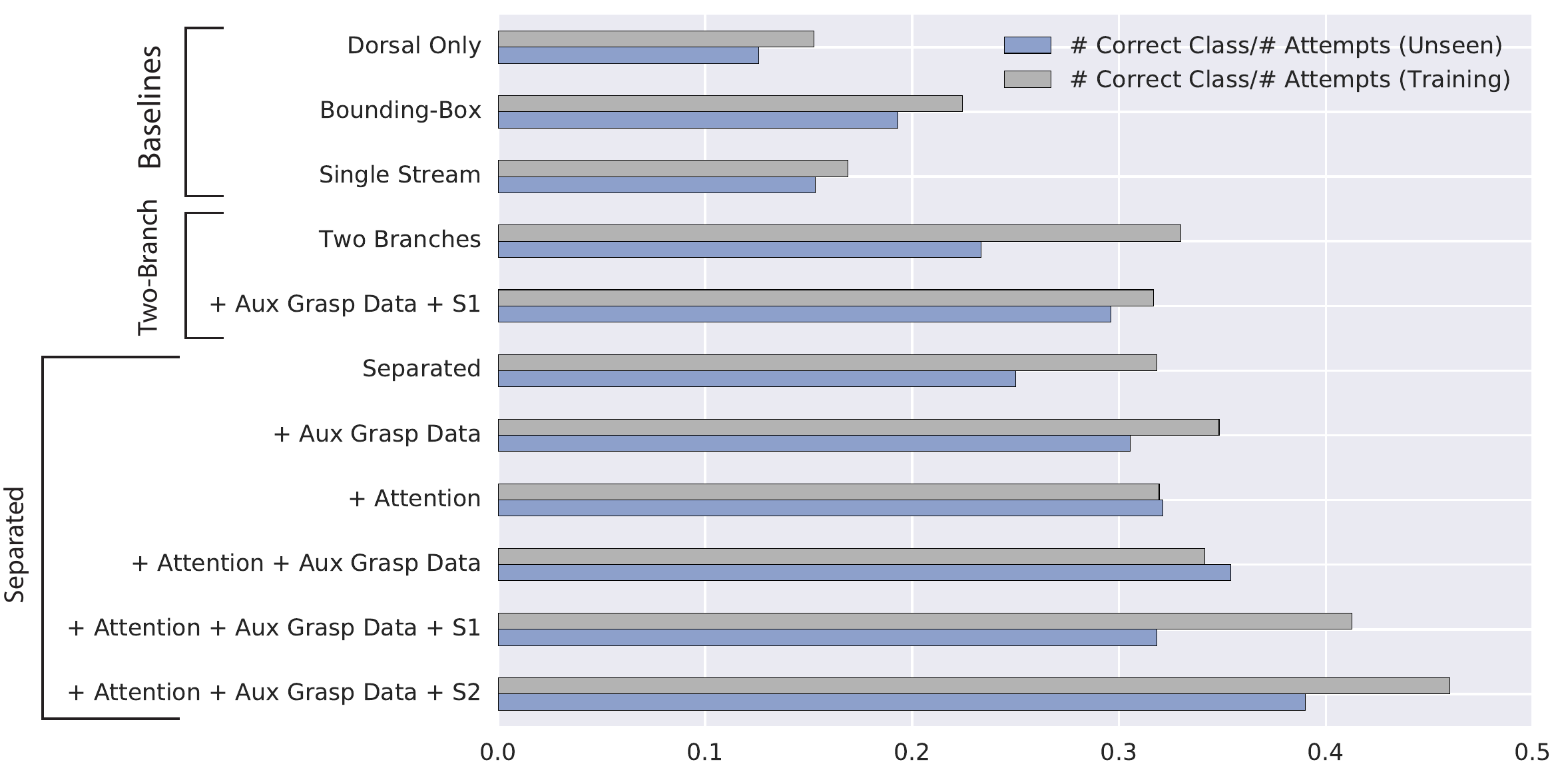}
\caption{Real-world robotic evaluation of semantic grasping models on training and unseen test objects, showing the percentage of grasp attempts that result in the correct class being grasped. See Appendices~\ref{sec:results_table},~\ref{sec:cms} for a complete table of evaluation metrics and class confusion matrices per model.\textsuperscript{2}}
\label{fig:barplot}
\end{figure}

\subsection{Baseline Comparisons}

The primary challenge in moving from non-semantic grasping to semantic grasping is that the policy must now extract a specific object from clutter, instead of just reaching for the object that is easiest to grasp. Furthermore, the actions that result in good grasps on a target object may not be the best actions that suggest the object's semantic identity (e.g., a handle on a shovel is easy to grasp but could also be the handle for a toilet brush). Our ``Dorsal Only'' baseline lacks semantic reasoning capabilities; while it can pick up an object 73.1\% and 73.5\% of the time on training and test objects, the fraction of successful grasps being of the right class is random (20.8\% on training objects and 17.2\% on test objects, Table~\ref{table:results}, ``Dorsal Only'') \footnote{We compute this via $.208 = .153 / .735$ and $.172 = .126 / .731$, see Appendix~\ref{sec:results_table}}.

\begin{wrapfigure}{r}{0.35\textwidth}
\centering
\includegraphics[width=.3\textwidth]{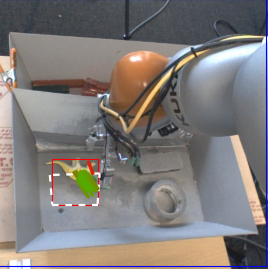}
\caption{Bounding box baseline.}
\label{fig:bbox}
\end{wrapfigure}

We compare our end-to-end semantic grasping policies to a semantic grasping approach in which object detection, classification, and grasp planning are separated into separate pipelines, which we refer to as the ``Bounding Box Model''. Since our grasping dataset lacks annotated segmentation masks for training region proposal models, we cannot train an object detector or classifier on grasping images; instead, we train an Inception-Resnet-V2 classifier~\cite{inception-resnet-v2} on the S1 dataset (Section \ref{sec:transfer}) to obtain a classifier with 95\% accuracy on held-out ``present images'' of training objects and 35\% accuracy on ``present images'' test objects collected during the data-collection phase. For object detection, we adopt a simplified version of RCNN~\cite{girshick2014rcnn, pas2017grasp} that performs a grid search within the initial image $I_0$ to find the fixed-size crop which most likely corresponds to the target class, as predicted by the trained classifier. The crop bounds in screen space are then projected to the workspace plane using precise camera extrinsic measurements, forming a quad that bounds the detected object. The non-semantic dorsal stream described in Section \ref{sec:twostream_arch} is then instructed to grasp an object constrained to these quad bounds (Figure \ref{fig:bbox}).

The ``Bounding Box'' model has 41.6\% and 32.4\% class accuracy among successfully grasped objects,
which is a significant improvement over the random behavior of a non-semantic dorsal grasping model. However, this baseline still had lower classification performance than all of our end-to-end learned models ($> 51.7\%$). Failure modes of this model include (1) failing to center the object in the detected crop, causing a poor grasp location on the object and (2) including more than one object in the crop, which may result in the arm picking up another object close to the target. These errors are primarily caused by a large domain shift from canonical image classification datasets to grasping images, despite the classifier being trained on present images of the same objects. Although learning a region-proposal model would result in better canonical crops for grasping, acquiring these labels is a labor-intensive procedure. In contrast, our models exploit closed loop robotic interaction with visual feedback to circumvent the need to acquire manually-annotated region-proposal annotations, and uses learned hand-eye coordination rather than camera calibration.

\subsection{Two-Stream Models}

We trained and evaluated 8 two-stream models, with 2 using a Two-Branch architecture and 6 using Separated architecture. Separated architectures performed better than Two-Branch models that jointly predicted class probability and grasp success (Appendix~\ref{sec:results_table}, rows 4,5 vs. rows 6,10).

Figure~\ref{fig:model}a visualizes the value function learned by our two stream models. In the left image, the image is superimposed with the actions that correspond to the highest probabilities of picking up any object. On the right, the image is superimposed with actions which are color-coded by the most likely object class at that location. The colors roughly correspond to the locations of objects, showing our model's ability to jointly learn visual grounding of semantic concepts.

A shallower model with soft keypoint attention (Section \ref{sec:soft_attn}) yields about the same semantic grasping performance as a deeper convolutional model while using half as many parameters (Appendix~\ref{sec:results_table}, rows 6,7 vs. rows 8,9). The effect of auxiliary grasping data has similar effects on both attention and non-attention models, suggesting that attention allows us to obtain comparable accuracy with fewer parameters. We use ``Separated + Attention'' as the base model for further experiments with transfer learning from auxiliary semantic datasets.

\subsection{Effect of External Dataset Transfer}
\label{sec:transfer_experiments}

We first examine the effect of auxiliary grasping data that is used to provide additional supervision to the grasp branch of our Two Branch architecture. In our ``Separated'' architectures, we train both grasp and class branches from the Two Branch model but only use the latter as the ventral stream of the two-stream model. Even when the grasp branch is replaced by a separately-trained model and not used for grasp prediction at evaluation, the auxiliary data still helps to learn low-level sensorimotor representations that are useful for classifying semantic objects. The Two-Stream Separated + Aux Grasp data yields a 3.0\% improvement on Class/Attempts over the Two-Stream Separated model, and the corresponding attention-based model yields a 2.2\% improvement on training objects and 3.3\% improvement on unseen objects.

Next, we consider the effect of transfer from the S1 and S2 datasets described in Section~\ref{sec:transfer}. While our models do not seem to benefit from external domain transfer of ImageNet/JFT, our experiments suggest that training on the ``optimal'' auxiliary dataset does improve classification success: S1 improves classification performance on training objects only, while training jointly on present images of test objects (S2) improves grasping accuracy on both training/test objects. These mixed results on domain transfer suggest that either JFT/ImageNet do not provide much information about our random selection of evaluated training and test objects, or that a more powerful domain adaptation architecture is required to generalize recognition of our object categories from these classification datasets. From a practical standpoint, this transfer is still useful as it enables robots to learn about object semantics without having to grasp them. However, it remains an open question as to how we can perform more effective knowledge transfer from datasets like ImageNet.

Overall, the results in Table~\ref{table:results} point to two-stream architectures being better suited to semantic grasping than single-stream models or a pipelined approach using an object detector. The overall success rate of the methods still leaves room for improvement, which indicates the difficulty of semantic grasping in clutter. This task combines the physical challenge of picking up diverse and varied objects with the finegrained perception challenge of predicting which object will be picked up. We hope that future work will continue to improve on these results, and that end-to-end semantic grasping will emerge as a useful benchmark task for both robotic control and situated object detection.


\section{Discussion}
\label{sec:conclusion}

In this work we propose a learning-based system for visual robotic grasping that combines spatial and semantic reasoning into a single neural network policy. Our network is decomposed into two streams, with the ``ventral'' stream being selective for the commanded object class while the ``dorsal'' stream selects for good grasps, independent of object identity.

Semantic grasping is not only a compelling research area for studying the intersection of spatial and semantic reasoning, but is also an as-yet unsolved task of considerable practical value. Semantic grasping capabilities could enable a household robot to clean a room, perform industrial sorting tasks on food products, or pack an ever-changing inventory of products into boxes. In each of these situations, the robot is likely to encounter objects that it has never seen or grasped before, and must appropriately generalize both spatial and semantic understanding to handle object diversity.

More broadly, robotics present an unique opportunity for cheaply computing labels for large amounts of complex data via active perception. We address the difficult problem of assigning semantic labels to large amounts of cluttered data via label propagation: during data collection, the robotic arm isolates the grasped object so that an uncluttered view of the object may be captured. This can be done by either lifting up the object to the camera for closer inspection, or using an auxiliary wrist-mounted camera to view the contents of the gripper directly. Such ``present images'' are centered, uncluttered, and easy to classify, and the temporal coherence of a grasping sequence allows us to retroactively assign the resulting semantic label to an earlier cluttered grasping scene. Our setup enables us to collect additional present poses per grasped object (e.g. multiple orientations), though we leave this for future work.

\clearpage

\acknowledgments{We would like to thank Vincent Vanhoucke for organization and discussion, John-Michael Burke for assistance with data collection and robot hardware, Zbigniew Wojna for helpful advice on attention mechanisms, Mrinal Kalakrishnan, Matthew Kelcey, and Ian Wilkes for code reviews, and Colin Raffel for helpful comments and feedback on the paper.}

\bibliography{semantic_bib} 

\appendix

\section{Data Collection Details}

\begin{figure}[H]
\centering
\includegraphics[width=0.8\linewidth]{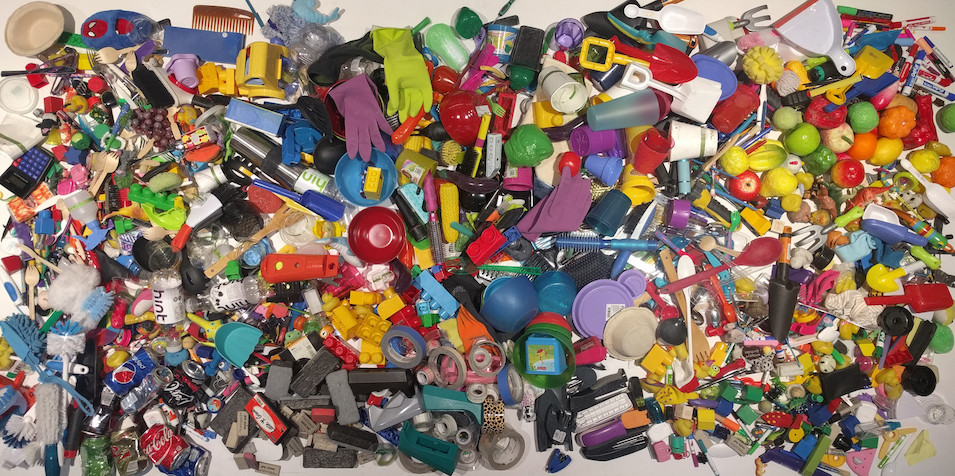}
\caption{A snapshot of all the objects used for data collection. The set contains 20-30 instances over 15-20 small objects that we might encounter in typical kitchen and office settings.}
\label{fig:training-objects}
\end{figure}

The Amazon Mechanical Turk service was used to label 20,000 present images for label propagation. The raters' choices included the $16$ categories in Table~\ref{table:object_classes} as well as \emph{No Objects}, \emph{Multiple}, and \emph{None of these}. We used a simple majority consensus to obtain the final label of each grasp sequence. Each task took under $20$ seconds and $2$ cents to complete on average. 

\section{Network Architectures}
\begin{figure}[H]
\centering
\includegraphics[width=\linewidth]{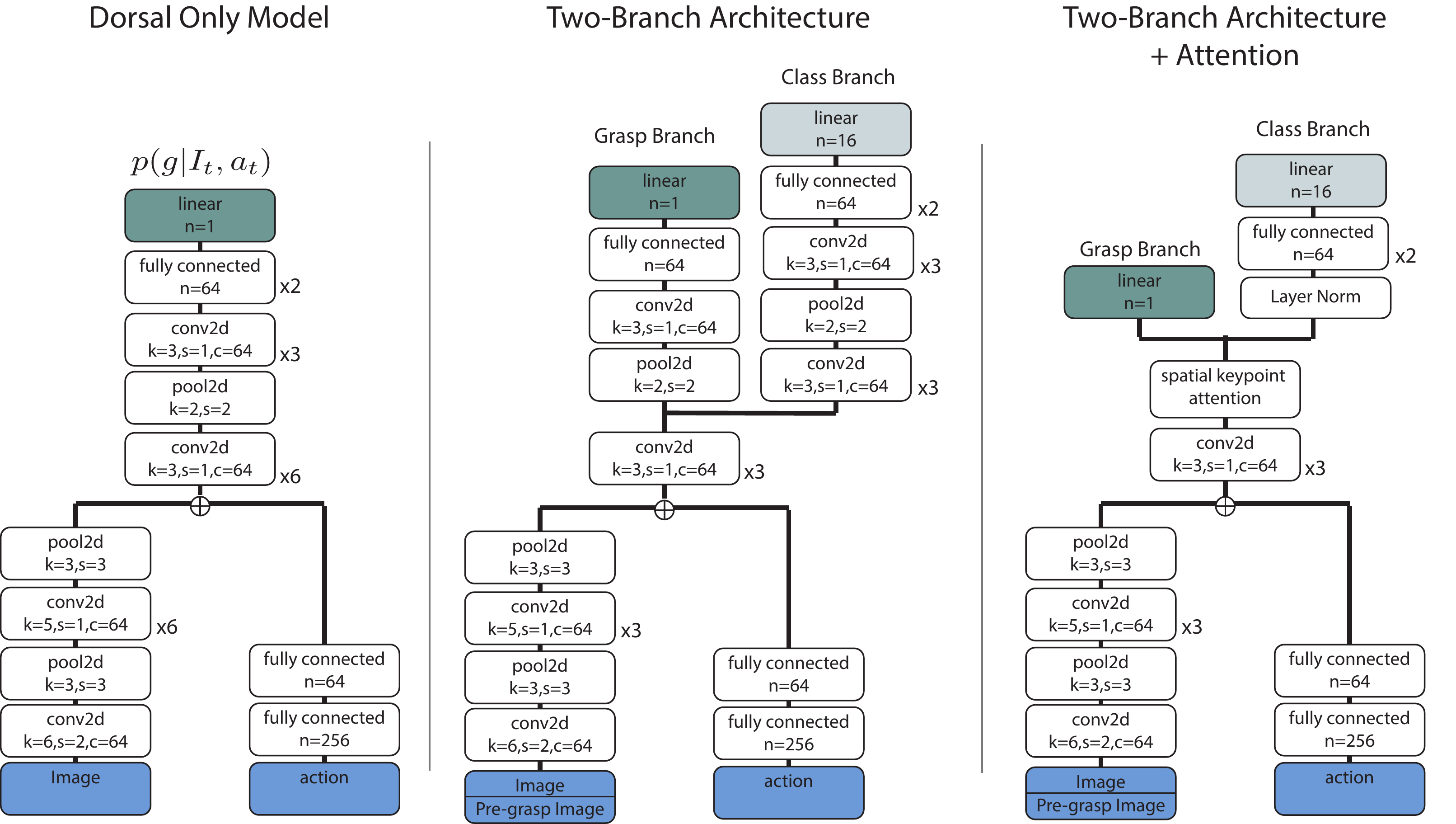}
\caption{Two-stream model for semantic grasping. ReLU activations and batch normalization ~\cite{is-bnad-15} are used at all but the final layer of each model.}
\label{fig:networkarch}
\end{figure}

\subsection{Soft Keypoint Attention}
\label{sec:soft_attn}

In soft keypoint attention \cite{dsae}, the output $a_{cij}$ of a convolutional layer is normalized across the spatial extent via a softmax $s_{cij} = e^{a_{cij} / \tau} / \sum_{i^\prime j^\prime} e^{a_{ci^\prime j^\prime}/ \tau}$. The spatial softmax for each channel is then used as weights for spatially averaging a mesh grid $f_c = (\sum_i i \cdot s_{cij}, \sum_j j \cdot s_{cij})$, which yields expected screen coordinates for salient visual features in that channel. These features are normalized to mean $0$ and standard deviation $1$ prior to classification, since class identity should be invariant to where the spatial features are located and their scale.

\begin{figure}[H]
\centering
\includegraphics[width=.3\textwidth]{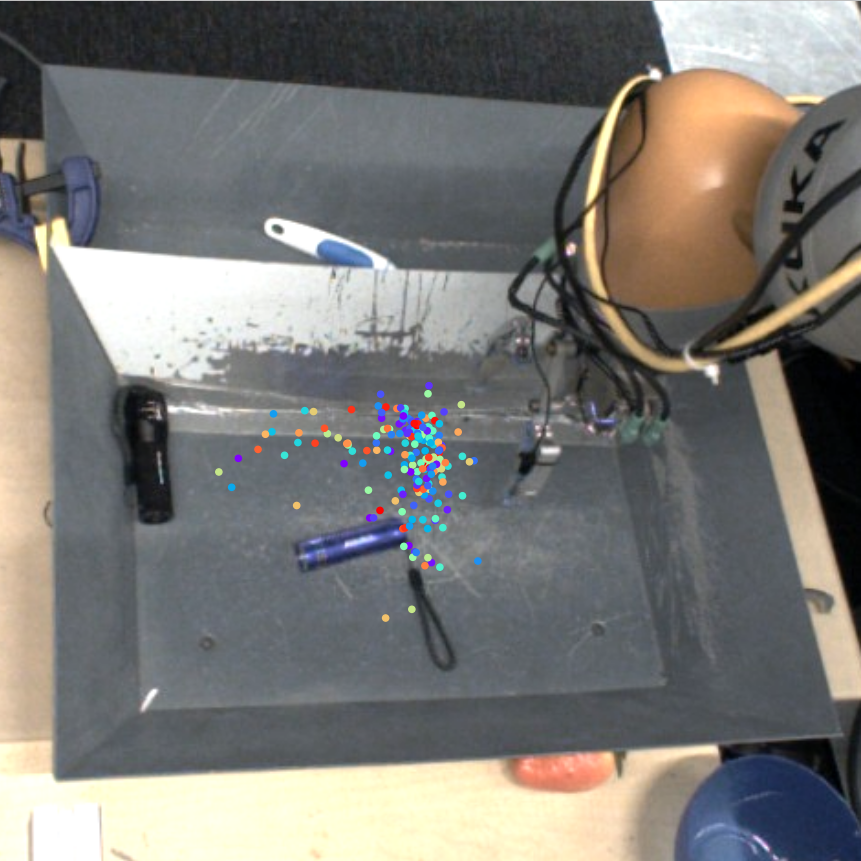}
\caption{Visualization of soft keypoint attention. Each point corresponds to the mean screen coordinates of a spatial softmax over channels in a convolution layer.}
\label{fig:soft_attn}
\end{figure}

Since soft keypoint attention returns a low-dimensional representation of attended features in the form of coordinates, we can directly model the dorsal and ventral branches of our model using a few fully-connected layers~\ref{fig:networkarch}. Although our models experimentally suggest that soft keypoint attention does not contribute to substantially better performance than simply using a deeper CNN, it is a convenient way to reduce the number of parameters in the model and accelerate training.

\section{Evaluation Details}
\label{sec:experiment_details}

Each grasp attempt consists of up to 10 time steps and at each time step, the robot records the current image $I_t$ and samples an action $a_t$ along which to move the gripper. At each step, action is optimized for several iterations via CEM~\cite{rk-cem-04}, which optimizes the neural critic described in Section~\ref{sec:twostream_arch} with respect to the proposed action. At the final grasp step $T$, the robot closes the gripper and evaluates the success of the grasp by recording feedback from the gripper. Data collection initially is bootstrapped on top of a policy that issues random motor commands at each time step. Off-policy data was collected $T=10$, with 2 iterations of CEM per step. At inference time, we used $T=4$ with 3 iterations of CEM per step to speed up the grasping system. The total numbers of grasps vary due to hardware failures that required restarting the evaluation protocol, resulting in some robots/objects having more attempts than others. However, grasps for a model were only scored after all evaluations were completed.

Occasionally, the robot picks up more than one object. If one of the objects grasped was the correct commanded one, we still count the grasp as a success (though the removal of two objects may result in another grasp failing).

Our semantic grasping models are trained using off-policy data from a non-semantic grasping policy, which tends to servo to the nearest object during data collection. Consequently, our dorsal and ventral streams are not trained on actions that are distal from the gripper, and thus cannot accurately reason about semantics and spatial reasoning for distant actions. Instead of the random approach move in \cite{levine16}, we modify the approach trajectory to hover the end-effector over 5 equally-spaced regions of the bin, so that the action search is able to consider moves from a variety of starting points (one of which will be close to the best candidate object). For each ``hover pose'', we optimize for the action that maximizes the semantic probability $p(c|I_t,a_t,g)$, and select the maximum-likelihood action among all such poses as the actual approach pose, in order to bring the gripper closer to the object.

Small probabilities from the ventral stream can destabilize predictions from the dorsal stream, even if the value function correctly identifies the right object to grasp. At inference time, we smooth the ventral probability via $\text{min}(p(c|I_t,a_t,g)+0.5, 1)$, prior to multiplying with the dorsal probability, which allows the robot to pick up an object even if it is unsure about the semantic identity of an object.

\section{Table of Results}
\label{sec:results_table}

\newcommand{\pbox}[1] {
  \parbox[c]{.85in}{
    \strut 
      #1
    \strut 
  }
}

\begin{table}[h]
  \centering
  \resizebox{1.\textwidth}{!}{
\begin{tabular}{|l|l|l|l|l|l|}
    \hline
      &   & \multicolumn{2}{|c|}{Training Objects} & \multicolumn{2}{|c|}{Unseen Objects} \\ \cline{2-6}
      & Model Name & {\scriptsize Objects/Attempts} & {\scriptsize Class/Attempts} & {\scriptsize Objects/Attempts} & {\scriptsize Class/Attempts} \\ 
    \hline 
    \multirow{2}{*}{\small Baselines}
      & {\small Dorsal Only} & 73.5 & 15.3 & 73.1  & 12.6 \\ \cline{2-6}
      & {\small Bounding Box} & 53.9 & 22.4 & 59.7  & 19.3 \\ \cline{2-6}   
      & {\small Single Stream} & 26.7 & 16.9 & 24.3  & 15.3 \\ \cline{2-6}
    \hline
    \multirow{2}{*}{\pbox{\small Two-Stream, Two-Branch}}
      & {\small Two Branches} & 50.3 & 33.0 & 43.0  & 23.3 \\ \cline{2-6}
      & \pbox{\small Two Branches + aux grasp data + S1} & 52.0 & 31.7 & 49.7  & 29.6 \\ \cline{2-6}
    \hline
    \multirow{2}{*}{\pbox{\small Two-Stream, Separated}}
      & {\small Separated} & 51.7 & 31.8 & 61.0 & 25.0 \\ \cline{2-6}
      & \pbox{\small Separated + Aux Grasp Data} & 62.5 & 34.8 & 56.4  & 30.5 \\  \cline{2-6}
      & \pbox{\small Separated + Attention} & 48.6 & 31.9 & 62.1  & 32.1 \\ \cline{2-6}
      & \pbox{\small Separated + Attention + Aux Grasp Data} & 51.1 & 34.2 & 63.0  & 35.4 \\ \cline{2-6}
      & \pbox{\small Separated + Attention + Aux Grasp Data + S1} & 62.0 & 41.2 & 66.9  & 31.8 \\ \cline{2-6}
      & \pbox{\small Separated + Attention + Aux Grasp Data + S2} & 64.0 & \textbf{46.0} & 61.7  & \textbf{39.0} \\ 
    \hline
\end{tabular}}
\caption{Objects/Attempts $=$ number of objects grasped divided by number of grasp attempts made (\%). Class/Attempts $=$ number of \textit{correct} objects grasped divided by number of grasp attempts (\%).}
\label{table:results}
\vspace{-0.3in}
\end{table}

\section{Confusion Matrices}
\label{sec:cms}

\begin{figure}[H]
\centering
\includegraphics[width=\linewidth]{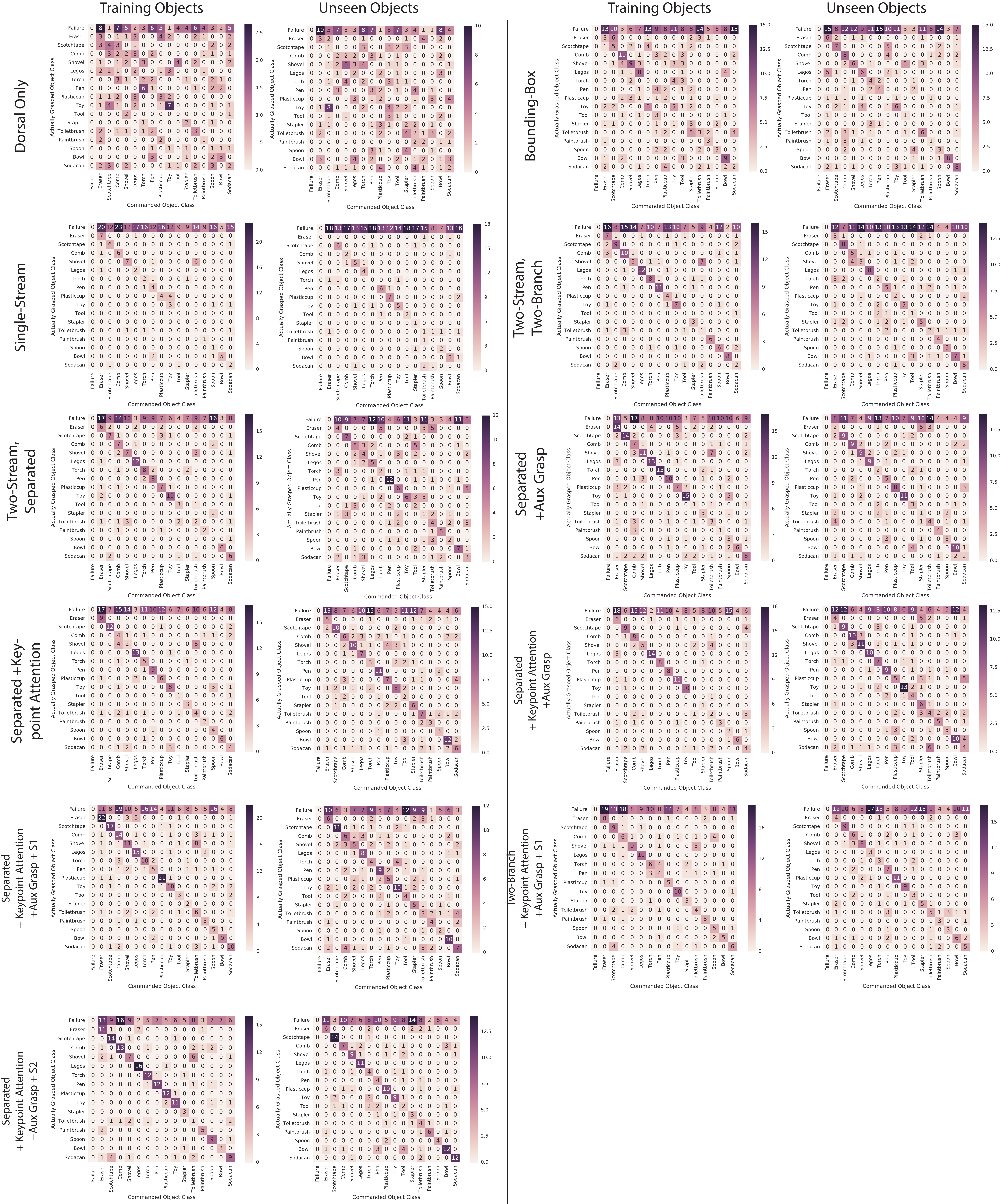}
\caption{Confusion matrices on grasping training and unseen objects for all evaluated models. Row $i$, column $j$ in each matrix corresponds to the number of times the robot picked up class $i$ when it was told to pick up class $j$.}
\label{fig:confusion}
\end{figure}

\end{document}